\documentclass[sigconf]{acmart}

\AtBeginDocument{%
  \providecommand\BibTeX{{%
    \normalfont B\kern-0.5em{\scshape i\kern-0.25em b}\kern-0.8em\TeX}}}

\let\labelindent\relax
\usepackage{enumitem}

\usepackage{tabularx}
\usepackage{booktabs} %
\usepackage{array}
\usepackage{subfigure}
\usepackage{listings}
\usepackage{xcolor}
\usepackage{mathrsfs}
\usepackage{fancyhdr}
\usepackage{multirow}
\usepackage{setspace}
\usepackage{colortbl}

\usepackage{pgfplots}
\pgfplotsset{compat=1.14}

\newcommand{\ignore}[1]{}
\usepackage{fancyhdr}
\usepackage[normalem]{ulem}
\usepackage{url}
\usepackage{microtype}
\usepackage{listings}
\usepackage{algorithm}
\usepackage{algpseudocode}

\usepackage{color}
\usepackage{xcolor}
\definecolor{Gray}{gray}{0.85}

\usepackage{xspace}

\makeatother

\usepackage[colorinlistoftodos,prependcaption,textsize=tiny]{todonotes}

\let\oldbibliography\thebibliography
\renewcommand{\thebibliography}[1]{%
  \oldbibliography{#1}%
  \setlength{\itemsep}{0pt}%
}

\usepackage[belowskip=4pt,aboveskip=4pt]{caption}

\usepackage[norndcorners,customcolors]{hf-tikz}
\hfsetbordercolor{yellow}
\hfsetfillcolor{yellow}
\definecolor{mygreen}{rgb}{0,0.6,0}
\definecolor{mygray}{rgb}{0.5,0.5,0.5}
\definecolor{mymauve}{rgb}{0.58,0,0.82}

\newcommand\blfootnote[1]{%
  \begingroup
  \renewcommand\thefootnote{}\footnote{#1}%
  \addtocounter{footnote}{0}%
  \endgroup
}

\newcommand{\theSystem}{CoDeNet}
\pgfplotsset{compat=1.16}

\copyrightyear{2021}
\acmYear{2021}
\setcopyright{acmlicensed}
\acmConference[FPGA '21] {2021 ACM/SIGDA International Symposium on Field Programmable Gate Arrays}{February 28--March 2, 2021}{Virtual Event, USA}
\acmBooktitle{2021 ACM/SIGDA International Symposium on Field Programmable Gate Arrays (FPGA '21), February 28--March 2, 2021, Virtual Event, USA}
\acmPrice{15.00}
\acmDOI{10.1145/3431920.3439295}
\acmISBN{978-1-4503-8218-2/21/02}
\begin{document}
\setlength{\parskip}{0mm}
\setlength{\dblfloatsep}{0.2cm}
\setlength{\dbltextfloatsep}{0.2cm}
\setlength{\floatsep}{0.2cm}
\setlength{\textfloatsep}{0.2cm}
\fancyhead{}

\title{CoDeNet: Efficient Deployment of Input-Adaptive Object Detection on Embedded FPGAs}
\author{Zhen Dong$^{*,1}$, Dequan Wang$^{*,1}$, Qijing Huang$^{*,1}$, Yizhao Gao$^2$, Yaohui Cai$^3$, Tian Li$^3$}
\author{Bichen Wu$^1$, Kurt Keutzer$^1$, John Wawrzynek$^1$}
\affiliation{
\textsuperscript{1}University of California, Berkeley, 
\textsuperscript{2}The University of Hong Kong, \textsuperscript{3}Peking University}   
\email{{zhendong, dqwang, qijing.huang, bichen, keutzer, johnw}@eecs.berkeley.edu}
\email{yizhao@connect.hku.hk,  {caiyaohui, davidli}@pku.edu.cn}
\renewcommand{\authors}{Qijing Huang, Dequan Wang, Zhen Dong, Yizhao Gao, Yaohui Cai, Tian Li, Bichen Wu, Kurt Keutzer, John Wawrzynek}
\begin{abstract}
\blfootnote{*Equal Contribution. \\
Archieved source code available at: \url{https://doi.org/10.5281/zenodo.4341394}} %
Deploying deep learning models on embedded systems for computer vision tasks has been challenging due to limited compute resources and strict energy budgets.
The majority of existing work focuses on accelerating image classification, 
while other fundamental vision problems, such as object detection, 
have not been adequately addressed. Compared with image classification, 
detection problems are more sensitive to the spatial variance of objects, 
and therefore, require specialized convolutions to aggregate spatial information. 
To address this need, recent work introduces dynamic deformable convolution to augment regular convolutions.
Regular convolutions process a fixed grid of pixels across all the spatial locations in an image, 
while dynamic deformable convolution may access arbitrary pixels in the image with the access pattern being input-dependent and varying with spatial location. 
These properties lead to inefficient memory accesses of inputs with existing hardware. 

In this work, we harness the flexibility of FPGAs to develop a novel object detection pipeline with deformable convolutions.
We show the speed-accuracy tradeoffs for a set of algorithm modifications including irregular-access versus limited-range and fixed-shape on a flexible hardware accelerator. 
We evaluate these algorithmic changes with corresponding hardware optimizations and show a 1.36$\times$ and 9.76$\times$ speedup respectively for the full and depthwise deformable convolution on hardware with minor accuracy loss. %
We then \textbf{Co}-\textbf{De}sign a \textbf{Net}work \textbf{\theSystem{}} with the modified deformable convolution for object detection and quantize the network to 4-bit weights and 8-bit activations. With our high-efficiency implementation, our solution reaches 26.9 frames per second with a tiny model size of 0.76 MB while 
achieving 61.7 AP50 on the standard object detection dataset, Pascal VOC. 
With our higher-accuracy implementation, our model gets to 67.1 AP50 on Pascal VOC with only 2.9 MB of parameters---$20.9\times$ smaller but 10\% more accurate than Tiny-YOLO.

\end{abstract}

\maketitle
\pagestyle{plain}
\section{Introduction}
Convolution is widely adopted in different neural network architecture designs for various object recognition tasks. 
Many hardware accelerators %
have been developed to improve the speed and power performance of the compute-intensive convolutional kernels. 
While the use of convolution kernels for computer vision is well-established,
researchers have been constantly proposing new operations and new network designs,
to increase the model capability and achieve better speed-accuracy trade-off for various tasks.
Deformable convolution~\cite{dai2017deformable, zhu2019deformable} is one of the novel operations that leads to state-of-the-art accuracy for object recognition with more effective use of parameters.
Many neural network designs with top accuracy~\cite{qiao2020detectors, zhang2020resnest} for object detection on the COCO dataset~\cite{lin2014microsoft} use deformable convolution in their design. 
Differing from conventional convolutions with fixed geometric structure, 
deformable convolution is an input-adaptive operation that samples inputs from variable offsets generated based on the input features during inference. 
Compared to conventional convolutions, deformable convolution provides a performance advantage due to: \textit{variable sampling scales} and \textit{variable sampling geometry}. 
The range for sampling at each different point varies, allowing the network to capture objects of different scales. 
Also, the geometry of the sample points is not fixed, allowing the network to capture objects of different shapes. 
Several previous studies~\cite{liu2018path}\cite{chen2019simpledet}\cite{li2019scale}\cite{zhou2019objects} have also shown that deformable convolution design lies on the Pareto-frontier of the speed-accuracy tradeoff for object detection on GPUs.

There are several challenges in supporting deformable convolution on off-the-shelf embedded deep learning accelerators: 
(i) The memory accesses for the input feature maps are irregular as they depend on the dynamically-generated offsets.
Many existing accelerators' instruction set architecture and the control logic are insufficient in supporting the random memory access patterns. 
In addition, the less contiguous memory access patterns limit the length of bursting memory accesses and incur more memory requests.   
(ii) There is less spatial reuse for the input features. 
Many accelerators are designed for output-stationary or row-stationary dataflow which leverages input reuse.
With deformable convolution, due to the variable filter offsets, the loaded input pixel for the current output pixel can no longer be reused by its neighboring output pixels. 
The lack of reuse significantly affects performance.
(iii) There is an increased memory bandwidth requirement for loading the variable offsets. %

FPGAs are well established to be ideal platforms for running object recognition tasks at the edge due to their power efficiency and low-batch inference performance.
Furthermore, timely and efficient hardware support for novel operations can be developed on FPGAs in weeks with high-level design tools.
For this work, we leverage the efficiency and flexibility of FPGAs, and available high-level tools, by adopting an algorithm-hardware co-design approach to address the challenges of efficient implementations of deformable convolutions.
We develop FPGA accelerators tailored to each algorithmic change and use these to study the accuracy-efficiency tradeoffs.%

We propose the following modifications to the deformable convolution operation to make it more hardware friendly: 
\begin{enumerate}%
	\item Limit the adaptive offsets to a fixed range to allow buffering of inputs and exploit full input reuse.
	\item Constrain the arbitrary offset displacements into a square shape to reduce the overhead from loading the offsets and to enable parallel accesses to on-chip memory.
    \item Round the offset displacements to integers and remove the fractional, bilinear interpolation operation for calculating the final sampling value.
    \item Use depth-wise convolution to reduce the total number of Multiply-Accumulate operations (MACs).
\end{enumerate}

We evaluate each modification on an FPGA System-on-Chip (SoC) that includes both an FPGA fabric and a hardened CPU core. 
We leverage the shared last-level cache (LLC) included in its full hardened processor system to efficiently exploit the locality of deformable convolution with data-dependent memory access patterns. 
We then optimize the hardware based on each algorithm modification to demonstrate its advantage in efficiency over the original operation. 
With these proposed algorithm modifications, 
we devise a line-buffer design to efficiently support our optimized depthwise deformable convolutional operation. 
To demonstrate the full capability of the co-designed operation, we also design an efficient deep neural network (DNN) model \theSystem{} for object detection using ShuffleNetV2~\cite{ma2018shufflenet} as the feature extractor. We quantize the network to 4-bit weights and 8-bit activations with a symmetric uniform quantizer using the block-wise quantization-aware fine-tuning process~\cite{dong2019hawq}. Our main contributions include: 
\begin{enumerate}%
    \item Co-design of a deformable convolution operation on FPGA with hardware-friendly modifications (depthwise, rounded-offset, limited-range, limited shape), showing up to  $9.76\times$ hardware speedup.
    \item Development of an efficient DNN model for object detection with codesigned input-adaptive deformable convolution that achieves 67.1 AP50 on Pascal VOC with 2.9 MB parameters. The model is $20.9\times$ smaller but $10\%$ more accurate than the Tiny-YOLO.
    \item Implementation of an FPGA accelerator to support the target neural network design that runs at 26 frames per second on Pascal VOC with 61.7 AP50.  
\end{enumerate}
The rest of the paper is organized as follows: Section~\ref{section:bg}
gives an introduction to the deformable convolution; Section~\ref{section:deform_conv} provides an ablation study for the operation and hardware co-design; Section~\ref{section:system} describes the end-to-end object detection system we design with the modified operation; Section~\ref{section:result} shows our final performance results; and we conclude the paper in Section~\ref{section:conclusion}.

\section{Background}
\label{section:bg}
\begin{figure}[t]
\centering
	\includegraphics[width=0.95\linewidth]{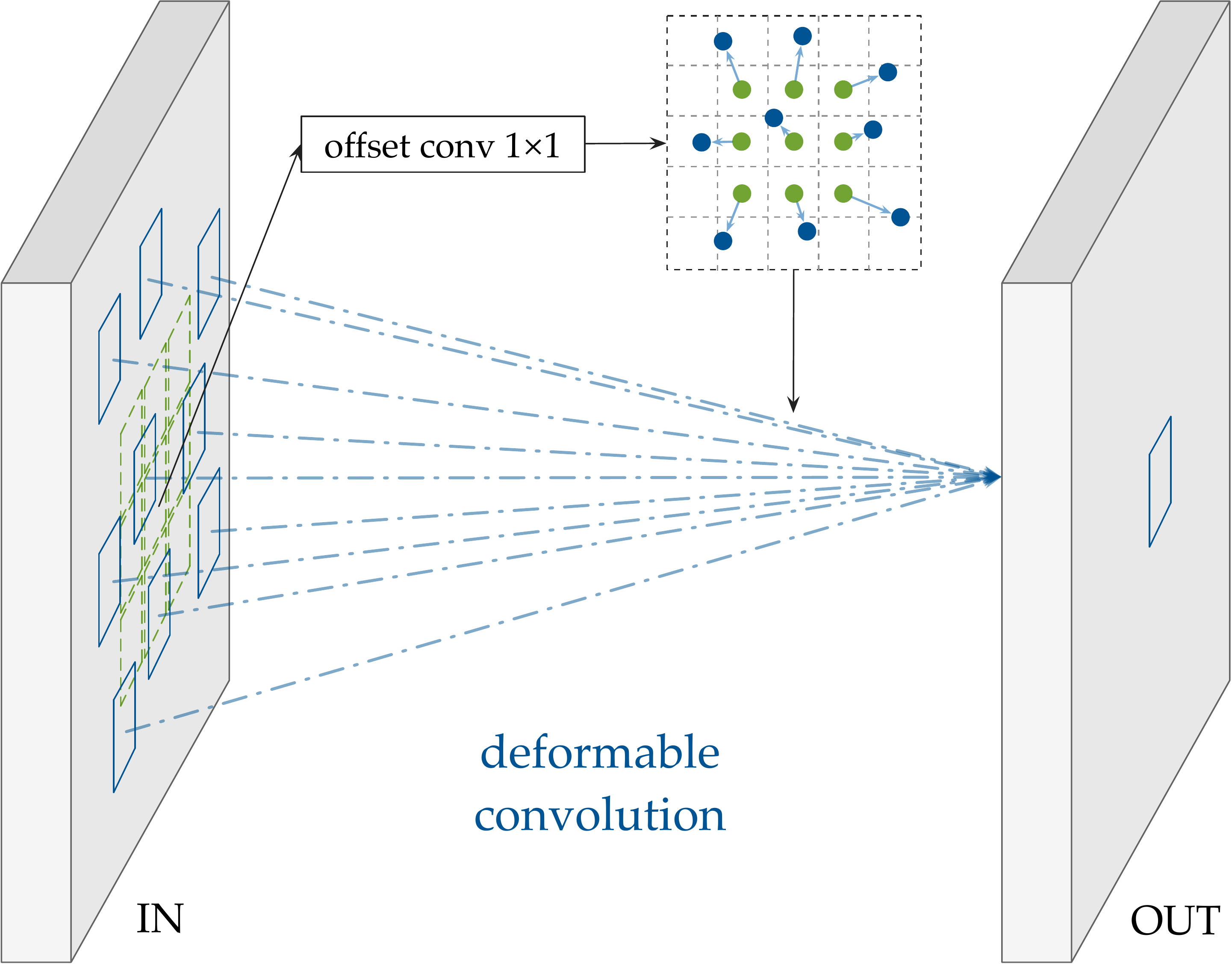}
\caption{Deformable convolution with input-adaptive displacement offsets generation. Deformable convolution in our design first generates the sampling offsets from the input feature map a using a 1$\times$1 convolution. Then it samples the same input feature map based on the generated offsets and performs a 3$\times$3 convolution to aggregate the corresponding spatial features. %
}
\label{fig:deform_op}
\end{figure}  
\subsection{Object Detection}
\label{section:2_1}
Object detection is a more challenging task than image classification as it performs object localization in addition to object classification and requires prediction on spatially-variant objects. %
Existing solutions for object detection can be categorized into two approaches: two-stage detector and one-stage detector.
In two-stage algorithms, the detector needs to first propose a set of regions of interest and then perform object classification on the selected regions.
Faster R-CNN~\cite{ren2015faster}, a two-stage algorithm, introduces Region Proposal Network (RPN) for efficient region proposal. 
RPN is widely adopted in two-stage algorithms as it reduces the overhead of region proposal by sharing features from the main detection network.
One-stage algorithms, on the other hand, skip the region proposal stage and directly run detection over a dense sampling of all possible regions.
Single Shot MultiBox Detector (SSD)~\cite{liu2016ssd}, a popular one-stage detector, leverages a pyramidal feature hierarchy in the feature extraction network to efficiently encode objects in various sizes. 
You Only Look Once (YOLO)~\cite{redmon2016you}\cite{redmon2017yolo9000} is another popular one-stage detector using fully convolutional network. 
The algorithm divides the input image into a grid with a fixed number of cells. 
Each cell in the grid predicts the bounding boxes of objects.
A prediction of the bounding box comprises location information, confidence scores, and the conditional probability of the object class. 
The location information consists of the coordinates of the object center and the object size. 
The confidence scores indicate the probability of an object in these boxes.

In this work, we use a one-stage anchor-free detector called CenterNet~\cite{zhou2019objects} due to its better Pareto efficiency for the speed-accuracy tradeoff compared to the concurrent works~\cite{duan2019centernet}\cite{law2018cornernet}\cite{law2019cornernet}\cite{zhou2019bottom}. 
In contrast to most anchor-free detectors where Non Maximum Suppression (NMS) mechanism is still required to remove the duplicated predictions, CenterNet directly generates the center points for each object without any post-processing. This property greatly reduces the complexity of implementing the detector pipeline in hardware. 

As for the evaluation metrics for object detection, a common practice is to use the average precision (AP) and intersection over union (IoU). %
AP computes the average precision value achieved with different recall values. 
The precision value, calculated as \\
$\frac{\text{true positive}}{\text{true positive} + \text{false positive}}$, indicates the percentage of predictions that are correct. 
The recall value, defined as $\frac{\text{true positive}}{\text{true positive} + \text{false negative}}$, measures the capability to correctly classify all positives. 
IoU is defined as the intersection between the predicted boxes and the target boxes over the union of the two.
The default evaluation metric for VOC dataset~\cite{everingham2010pascal} is AP50, which indicates that the prediction would be seen as correct if the corresponding IoU $\geq 0.5$.
The main metric for COCO is the mean of the average precisions at IoU from 0.5 to 0.95 with a step size of 0.05.

\subsection{Deformable Convolution}
\begin{figure}[!t]
\begin{minipage}[t]{0.48\linewidth}
    \centering
	\includegraphics[trim={4cm 1cm 1cm 1cm},clip, width=\linewidth]{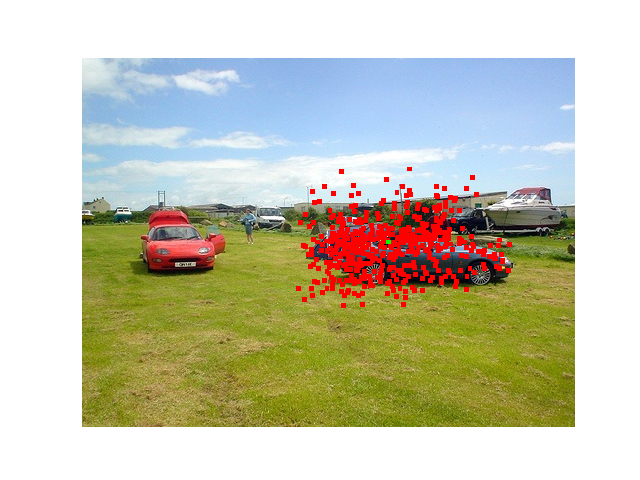}
	\caption*{(a) car}
\end{minipage}
\begin{minipage}[t]{0.48\linewidth}
    \centering
	\includegraphics[trim={4cm 1cm 1cm 1cm},clip,width=\linewidth]{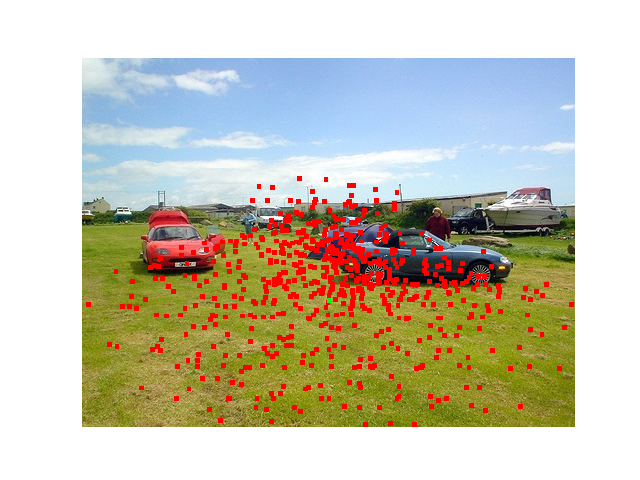}
	\caption*{(b) lawn}
\end{minipage}

\caption{Example for the input-adaptive deformable convolution sampling locations and offset range distribution for different active detection units. (a) the sampling locations for the car as an active unit. (b) the sampling locations for lawn in the background. }%
\label{fig:sample_example}
\vspace{-5pt}
\end{figure}  

Compared to image classification,
one challenge in object detection is to capture geometric variations of each object,
such as scale, pose, viewpoint, and part deformation.
Besides, different objects located in different regions of the same image can be geometrically different, making it hard to capture all features in one pass. 
State-of-the-art approaches~\cite{chen2019simpledet}\cite{li2019scale}\cite{liu2018path}\cite{shelhamer2019blurring}\cite{zhou2019objects} address these challenges by harnessing deformable convolution~\cite{dai2017deformable}\cite{zhu2019deformable}. 
As demonstrated in Figure~\ref{fig:deform_op},
deformable convolution samples the input feature map using the offsets dynamically predicted from the same input feature map, after which it performs a regular convolution over the features sampled from the predicted offsets. 
The convolution layer for generating the offsets
is typically composed of one 1$\times$1 or 3$\times$3 convolution layer. It is jointly trained with the rest of the network using standard backpropagation in an end-to-end manner. 
This way the gradient updates not only the weights of the convolutions but also the sampling locations for the convolutions.
Such operation design enables more flexible and adaptive sampling on different input feature maps. 

Unlike the regular convolution with fixed geometry, 
the receptive fields of deformable convolution can be of various shapes to capture objects with different scales, aspect ratios, and rotation angles. 
In addition, deformable convolution is both spatial-variant and input-adaptive. In other words, 
its sampling patterns and offsets vary for different output pixels in the same input feature map and also vary across different input feature maps.
In Figure~\ref{fig:sample_example}(a)(b), we show how the sampling locations (red dots) change with the different active detection units (the object with a green dot on it). %
Most of the offsets are within the $[-1,4]$ range for the example image. 
Albeit the operation augments and enhances the capability of the existing convolution for object detection, its dynamic nature poses extra challenges to the existing hardware. %

\subsection{Algorithm-hardware Co-design for Object Detection}
Many prior acceleration works~\cite{zhu2018euphrates}~\cite{ma2018algorithm}\cite{nakahara2018lightweight}\cite{hao2019fpga}\cite{zhang2019skynet}\cite{xu2019scalable}~\cite{wijesinghe2019hardware} have demonstrated the effectiveness of the co-design methodology for the deployment of real-time object detection on FPGAs. 
~\cite{ma2018algorithm} customizes SSD300~\cite{liu2016ssd} by replacing operations, such as dilated convolutions, normalization, and convolutions with larger stride, with more efficiently supported ones on FPGAs. 
~\cite{nakahara2018lightweight} adapts YOLOv2~\cite{redmon2017yolo9000} by introducing a binarized network as the backbone for feature extraction to leverage the low-precision support of FPGA. 
Meanwhile, the FINN-R framework ~\cite{blott2018finn} further explores the benefits of integrating quantized neural networks (QNN) into Yolo-based object detection systems. 
Real-time object detection for live video streaming system~\cite{preusser2018inference} enables is then developed with the FINN-based QNNs.
~\cite{hao2019fpga} devised an automatic co-design flow on embedded FPGAs for the DJI-UAV~\cite{xu2019dac} dataset with 95 categories targeting unmanned aerial vehicles.  %
The flow first constructs DNN basic building blocks called bundles, estimates their corresponding latency and cost on hardware, and selects the ones on the pareto front for latency and resources trade-off. Then it starts a two-phase DNN evaluation to search for the bundles on the pareto front of the accuracy-latency trade-off and then fine-tune the design of the selected bundles.    
SkyNet~\cite{zhang2019skynet} searched by this co-design flow achieves the best performance (based on a combination of throughput,
power, and detection accuracy) on embedded GPUs and FPGAs.
Differing from prior work, we study a novel and efficient operation, deformable convolution, for object detection. 
In addition to modifying the neural network design, we also co-design the operation for better hardware efficiency. 

\begin{figure*}[t]
	\centering
	\subfigure[normal]{
	\begin{minipage}[t]{0.18\linewidth}
		\centering
		\includegraphics[page=1, width=\linewidth]{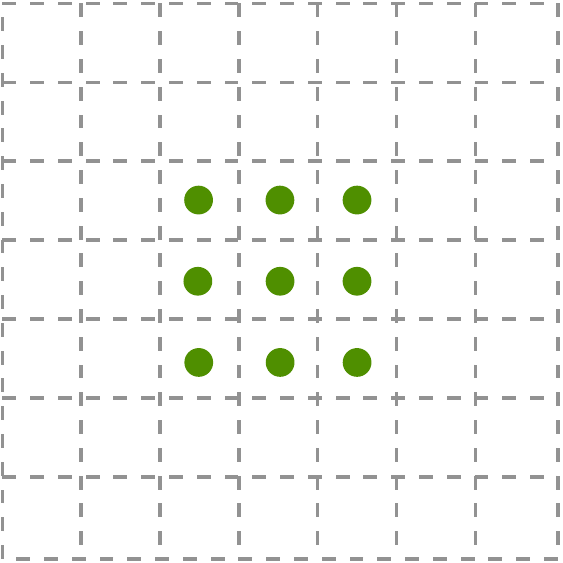}
	\end{minipage}
	}
	\subfigure[deform]{
	\begin{minipage}[t]{0.18\linewidth} 
		\centering
		\includegraphics[page=2, width=\linewidth]{images/deform_conv.pdf}
	\end{minipage} 
	}
	\subfigure[bound]{
	\begin{minipage}[t]{0.18\linewidth} 
		\centering
		\includegraphics[page=3, width=\linewidth]{images/deform_conv.pdf}
	\end{minipage} 
	}
	\subfigure[square]{
	\begin{minipage}[t]{0.18\linewidth}
		\centering
		\includegraphics[page=4, width=\linewidth]{images/deform_conv.pdf}
	\end{minipage}
	}
	\subfigure[round]{
	\begin{minipage}[t]{0.18\linewidth} 
		\centering
		\includegraphics[page=5, width=\linewidth]{images/deform_conv.pdf}
	 \end{minipage}  
	}
	\caption{Major algorithm modifications for deformable convolution operational co-design. (a) is the default 3$\times$3 convolutional filter. (b) is the original deformable convolution with unconstrained non-integer offsets. (c) sets an upper bound to the offsets. (d) limits the geometry to a square shape. (e) shows that the predicted offsets are rounded to integers.}
\end{figure*}

\subsection{Quantization}
\label{section:2_4}
Quantization~\cite{zhou2016dorefa}\cite{jacob2018quantization}\cite{zhang2018lq}\cite{dong2019hawq}\cite{cai2020zeroq} is a critical technique for efficiently deploying neural network models on embedded devices. It alleviates the memory bottleneck by compressing the weights in neural network models into ultra-low precision such as 4 bits. Moreover, quantizing both the weights and activations enables the use of cheaper low-precision integer arithmetics on hardware.
For DNN deployment on embedded FPGAs without floating-point arithmetic support, quantization is one key and necessary modification.

However, directly performing aggressive layer-wise quantization can result in significant accuracy degradation~\cite{krishnamoorthi2018whitepaper}. Many prior works have attempted to address this accuracy drop with various techniques, such as non-uniform learnable quantizer~\cite{zhang2018lq}, mixed-precision quantization~\cite{dong2019hawqv2}, progressive fine-tuning~\cite{zhou2017incremental} as well as group-wise~\cite{shen2019q} and channel-wise quantization~\cite{krishnamoorthi2018whitepaper}. 
Although these methods can better preserve the accuracy of the pre-trained model, they increase the complexity of hardware implementation and can introduce non-negligible overhead on both latency and memory usage. Consequently, it is crucial to carefully consider the trade-off between accuracy and hardware efficiency when quantizing a model for the edge devices.
Quality of quantization is also strongly correlated to the network architecture and the target task. ~\cite{krishnamoorthi2018whitepaper} shows that compact models are more difficult to quantize. 
Besides, compared to image classification, object detection is a more challenging task for ultra-low precision quantization because it requires accurate localization of specific objects in an image.
Even with quantization-aware fine-tuning, quantizing the detection models with naive quantization schemes can cause around $10\%$ AP degradation on the COCO dataset~\cite{li2019fully}. In this work, we take advantage of mixed-precision quantization where we have 4-bit for weights and 8-bit for activations. This can significantly reduce the accuracy degradation since activations are more sensitive compared to weights in object detectors.

\section{Deformable Operation Co-design}
\label{section:deform_conv}

Although deformable convolution augments the neural network design with input-adaptive sampling, it is challenging to provide efficient support for the operation in its original form on hardware accelerators due to the following reasons:
\begin{enumerate}
    \item the limited reuse of input features
    \item the irregular input-dependent memory access patterns 
    \item the computation overhead from the bilinear interpolation
    \item the memory overhead of the deformable offsets
\end{enumerate}
In this work, we perform a series of modifications to deformable convolution with the objective to enable more data reuse and higher degree of parallelism for FPGA acceleration.  
A comprehensive ablation study is done to demonstrate the impact of each algorithmic modification on accuracy. 
We perform our study with standard object detection benchmarks, VOC and COCO.
We then design a specialized hardware engine optimized for each algorithmic modification on FPGA and show the performance improvement on FPGA from each modification. 
The accuracy and hardware efficiency trade-off is studied for each modification we propose.

We will be using the following notations in the paper: 
$n$ - batch size, $h$ - height, $w$ - width, $ic$ - input channel size, $oc$ - output channel size, $k$ - kernel size, $\Delta p$ - offsets.

\subsection{Algorithm Modifications}
\label{section:deform_conv_algorithm}
\begin{table*}[!t]
    \centering
    \caption{Ablation study of operation choices for object detection on VOC and COCO. The top half shows the baselines with various kernel sizes, from 3$\times$3 to 9$\times$9. The bottom half shows the comparison of different designs for deformable convolution.}
    \resizebox{0.9\textwidth}{!}{
    \begin{tabular}{cccc|ccc|ccc|ccc}
    \hline
        \multirow{2}{*}{\textbf{Operation}} & \multirow{2}{*}{\textbf{Depthwise}} & \multirow{2}{*}{\textbf{Bound}} & \multirow{2}{*}{\textbf{Square}} & \multicolumn{3}{c|}{ \textbf{VOC}} & \multicolumn{6}{c}{ \textbf{COCO}} \\ \cline{5-13}
         &  &  &  & \textbf{AP} & \textbf{AP50} & \textbf{AP75} & \textbf{AP} & \textbf{AP50} & \textbf{AP75} & \textbf{APs} & \textbf{APm} & \textbf{APl} \\ 
    \hline
        $3\times3$ & & & & 39.2 & 60.8 & 41.2 & 21.4 & 36.5 & 21.5 & 7.3 & 24.1 & 33.0 \\
        $3\times3$ & \checkmark & & & 39.1 & 60.9 & 40.9 & 19.8 & 34.3 & 19.7 & 6.3 & 22.6 & 31.5\\
        $5\times5$ & \checkmark & & & 40.6 & 62.4 & 42.6 & 21.3 & 36.4 & 21.3 & 6.7 & 23.7 & 34.2\\
        $7\times7$ & \checkmark & & & 41.9 & 63.8 & 43.8 & 21.7 & 37.2 & 21.5 & 6.9 & 24.0 & 35.2\\
        $9\times9$ & \checkmark & & & 42.3 & 64.8 & 44.3 & 22.2 & 37.8 & 22.1 & 7.0 & 24.3 & 35.4\\
    \hline
        deform & \checkmark & & & 42.9 & 64.4 & 45.7 & 23.0 & 38.4 & 23.3 & 6.9 & 24.4 & 37.8 \\
        deform & \checkmark & \checkmark & & 41.0 & 63.0 & 42.9 & 21.3 & 36.4 & 21.1 & 7.2 & 23.6 & 34.4 \\
        deform & \checkmark & \checkmark & \checkmark & 41.1 & 63.1 & 43.7 & 21.5 & 36.8 & 21.5 & 6.5 & 23.7 & 34.8 \\
    \hline
    \end{tabular}}
\label{tab:algo_codesign}
\end{table*}

We choose average precision (AP) as the main metric for benchmarking object detection performance on VOC and COCO datasets. 
ShuffleNet V2~\cite{ma2018shufflenet} is used as the feature extractor in all experiments.
As for decoder, we follow the practice of CenterNet~\cite{zhou2019objects} and use the stack of deformable convolution, nearest $2\times$ upsample, and ReLU activation layers.
Table~\ref{tab:algo_codesign} lists the modifications we make to the original deformable convolution as well as a comparison among deformable convolutions of different forms and regular convolutions with different kernel sizes. 
From the comparison, we see that the original deformable convolution achieves higher accuracy on Pascal VOC compared to convolution with $9\times9$ kernel (42.9 vs 42.3) while requiring $\frac{9\times9}{3\times3} = 9\times$ fewer MACs and weight parameters. %
Here we discuss how we further improve the efficiency of deformable convolution for hardware step-by-step.

\textbf{Depthwise Convolution}
We first replace the full 3$\times$3 deformable convolutions with 3$\times$3 depthwise deformable convolutions and 1$\times$1 convolutions,
similar to the depthwise separable convolution practice in Xception~\cite{chollet2017xception}. 
Such modification makes the whole network more uniform and smaller, so the weights of the deformable convolution can be all buffered on-chip for maximal reuse.

\textbf{Bounded Range}
Our next algorithmic modification to facilitate efficient hardware acceleration is to restrict the offsets to a positive range. %
Such constraint limits the size of the working set of feature maps 
so that a pre-defined fixed-size buffer can be added to the hardware,
in order to further exploit the temporal and spatial locality of the inputs. 
Assume a uniform distribution for the generated offsets in a $3\times3$ convolution kernel with stride $1$,
each pixel is expected to be used nine times. 
If all inputs within the range can be stored in the buffer, 
all except the first access to the same address will be from on-chip memory with $1 \sim 3$ cycle latency. 
We impose this constraint during training by adding a \textit{clipping} operation after the offset generation layer to truncate offsets that are smaller than 0 or larger than $N$, 
so all offsets $\Delta p_x, \Delta p_y \in [0, N]$.
Table~\ref{tab:algo_codesign} shows that 
setting the bound $N$ to 7 results in 1.9 and 1.7 AP degradation on VOC and COCO respectively.

\textbf{Square Shape}
Another obstacle to efficiently supporting the deformable convolution is its irregular data access patterns,
which leads to serialized memory accesses to multi-banked on-chip memory. 
To address this issue, we further constrain the offsets to be on the edges of a square. 
Instead of using $3\times3\times2 = 18$ numbers to represent the $\Delta p_x$ and $\Delta p_y$ offsets for all nine samples, only one number $\Delta p_d$, 
representing the distance from the center to the sides of the square, needs to be learned. 
This is similar to a dilated convolution with spatial-variant adaptive dilation factors. %
Adding this modification leads to 0.1 and 0.2 AP increase on VOC and COCO.

\textbf{Rounded Offsets}
In the original deformable design,
the generated offsets are typically fractional and a bilinear interpolation needs to be performed to produce the target sampling value. 
Bilinear interpolation calculates a weighted average of the neighboring pixels for a fractional offset based on its distance to the neighboring pixels. 
It introduces at least six multiplications to the sampling process of each input,
which is a significant increase ($6 \times h \times w \times ic $) to the total FLOPs.
We thus round the offsets to be integers during inference to reduce the total computation. 
The dynamically-generated offsets are thus rounded to integers. %
In practice, we round the generated offset during the quantization step.

As shown in Table~\ref{tab:algo_codesign}, together with the modifications above, our co-designed deformable convolution achieves 41.1 and 21.5 AP on VOC and COCO respectively, which is 1.8 and 1.5 lower than the original depthwise deformable convolution. 
Note that the accuracy of the modified deformable convolution still achieves higher accuracy compared to the large $5\times5$ kernel, while requiring $\frac{3\times3}{5\times5}=36\%$ fewer MACs and parameters.

\begin{table*}[!t]
\centering
\caption{Co-designed hardware performance comparison. The top half shows the performance of codesigned hardware corresponding to each algorithmic changes to the default 3$\times$3 convolution. The bottom half shows the results for the depthwise 3$\times$3 convolution.}
	\begin{tabular}{cccc|cccc}
	\hline
	\multirow{2}{*}{\textbf{\small Operation}} & \multirow{2}{*}{\textbf{\small Deform}} &  \multirow{2}{*}{\textbf{\small Bound}} &  \multirow{2}{*}{\textbf{\small Square}} & \multicolumn{2}{c}{\textbf{\small Without LLC}} & \multicolumn{2}{c}{\textbf{\small With LLC}} \\
	\cline{5-8}
	&  &  &  & \small{Latency (ms)} & \small{GOPs} & \small{Latency (ms)} & \small{GOPs} \\
	\hline
	&  & & & 43.1 & 112.0 & 41.6 & 116.2 \\
	default &  \checkmark & & & 59.0 & 81.8 & 42.7 & 113.1 \\
	3$\times$3 conv &  \checkmark & \checkmark & & 43.4 & 111.5 & 41.8 & 115.5 \\
	&  \checkmark & \checkmark & \checkmark & 43.4 & 111.5 & 41.8 & 115.6 \\
	\hline
	&  & & & 1.9 & 9.7 & 2.0 & 9.6 \\
	depthwise &  \checkmark & & & 20.5 & 0.9 & 17.8 & 1.1 \\
	3$\times$3 conv &  \checkmark & \checkmark & & 3.0 & 6.2 & 3.4 & 5.5 \\
	&  \checkmark & \checkmark & \checkmark & 2.1 & 9.2 & 2.3 & 8.2 \\
	\hline
	\end{tabular}
\label{tab:hw_codesign}
\end{table*}
\begin{figure}[t]
\centering
    	\includegraphics[width=\linewidth]{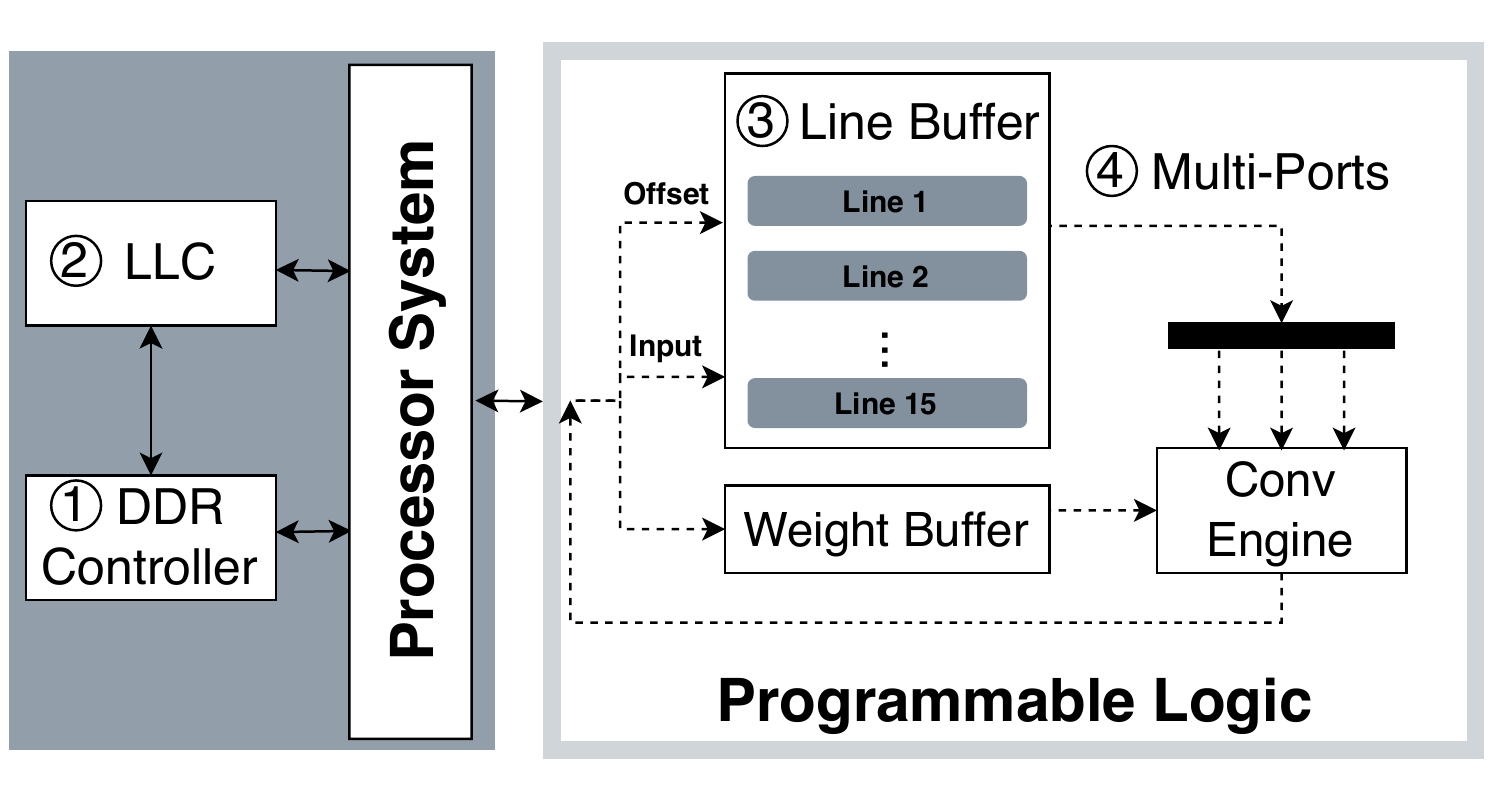}
\caption{Hardware engine for deformable convolution.} %
\label{fig:hw_codesign}
\end{figure} 
\subsection{Hardware Optimizations}
\label{section:3_2}
Many hardware optimization opportunities are exposed after we perform the aforementioned modifications to deformable convolution. We implement a hardware deformable convolution engine on FPGA SoC as shown in Figure~\ref{fig:hw_codesign} and tailor the hardware engine to each algorithm modification. The experiments are run on the Ultra96 board featuring a Xilinx Zynq XCZU3EG UltraScale+ MPSoC platform. The accelerator logic accesses the 1MB 16-way set-associative LLC through the Accelerator Coherency Port (ACP).  The data cache uses a pseudo-random replacement policy.  
Table~\ref{tab:hw_codesign} lists the speed and throughput performance for different customized hardware running a kernel of size $h=64, w=64, k=256, c=256$. 
In all experiments, we round the dynamically-generated offsets to integers. We use $8\times8\times9$ Multiply-Accumulate (MAC) units in the $3\times3$ convolution engine for all full convolution experiments and $16\times9$ MACs for depthwise convolution experiments.

\textbf{Baseline}
The baseline hardware implementation for the original $3\times3$ deformable convolution directly accesses the DRAM without going through any cache or buffering. In Figure~\ref{fig:hw_codesign}, the baseline implementation directly accesses the input and output data through HP ports and  \textcircled{1} DDR controller.
The input addresses are first calculated from the offsets loaded from DRAM. The \textit{$3\times3$ Deform M2S} engine then fetches and packs the inputs into parallel data streams to feed into the MAC units in the \textit{$3\times3$ Conv} engine. 
This baseline design resembles accelerator designs with only a scratchpad memory that cannot leverage the temporal locality of the dynamically loaded inputs for deformable convolution.

\textbf{Caching}
One hardware optimization to leverage the temporal and spatial locality of the nonuniform input accesses is to add a cache to the accelerator system. 
As shown in Figure~\ref{fig:hw_codesign}, we load the inputs from \textcircled{2} LLC through the ACP port in this implementation to reduce the memory access latency of the cached values.  
Since the inputs are sampled from offsets without specific patterns in the original deformable convolution, the cache provides adequate support to buffer inputs that might be reused in the near future. 
As shown in Table~\ref{tab:hw_codesign}, adding LLC results in 27.6\% and 13.2\% reduction in latency for the original full and depthwise deformable convolution respectively. 

\textbf{Buffering}
With the bounded range modification to the algorithm, we are able to use the on-chip memory to buffer all possible inputs. 
Similar to a line-buffer design for the original $3\times3$ convolution that stores two lines of inputs %
to exploit all input locality, we store $2N$ lines of inputs so that it is sufficient to buffer all possible inputs for reuse. This implementation includes the \textcircled{3} Line Buffer in Figure~\ref{fig:hw_codesign}.
With the effective buffering strategy, we can see in Table~\ref{tab:hw_codesign} that the latency of a bounded deformable is reduced by 26.4\% and 85.3\% for full and depthwise convolution respectively in a system without LLC. In a system with LLC, the reduction is 2.1\% and 80.9\% respectively.  The depthwise deformable convolution benefits more from adding the buffer as it is a more memory-bound operation. The compute-to-communication ratio for its input is $oc$ times lower than the full convolution.

\textbf{Parallel Ports}
The algorithm change to enforce a square-shape sampling pattern not only reduces the bandwidth requirements for loading the input indices in hardware, but also helps to improve the on-chip memory bandwidth.  
With a non-predictable memory access pattern to the on-chip memory, only one input can be loaded from the buffer at each cycle if all sampled inputs are store in the same line buffer. 
By constraining the shape of deformable convolution to a square with variable dilation,
we are guaranteed to have three different line buffers with each storing three sampled points.
We can thus have three parallel ports (\textcircled{4} Multi-ports in Figure~\ref{fig:hw_codesign}) accessing different line buffers concurrently. 
This co-optimization improves the on-chip memory bandwidth and leads to another $\sim$ 30\% reduction in latency for depthwise deformable convolution. 

With the co-design methodology, our final result shows a 1.36$\times$ and 9.76$\times$ speedup respectively for the full and depthwise deformable convolution on the embedded FPGA accelerator. 
These optimizations can also be beneficial to other hardware with line buffer and parallel ports support.

\section{Detection System Co-Design}
\label{section:system}
In addition to the deformable convolution operation, the design of feature extractor, detection heads and quantization strategy, also significantly impact the accuracy and efficiency of our detection system.
In this section, we introduce \theSystem{} for efficient detector and a specialized FPGA accelerator design to support it.

\begin{figure*}[!t]
    \centering
    \includegraphics[page=1, width=0.98\linewidth]{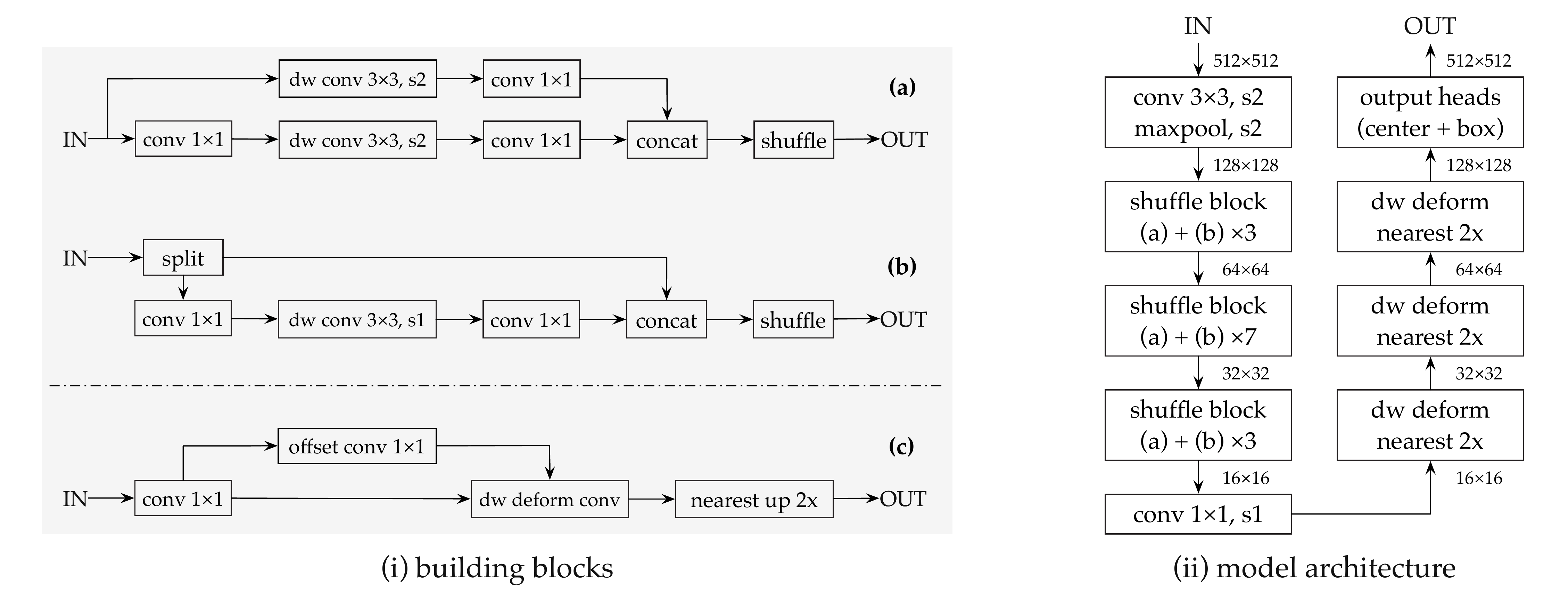} 
\caption{The architecture diagrams of our building blocks and model architecture.See section \ref{section:system_design} for more details.}
\label{fig:building_blocks}
\end{figure*} 

\begin{figure}[t]
	\centering
	\subfigure[image]{
	\begin{minipage}[t]{0.22\linewidth}
		\centering
		\includegraphics[page=1, width=\linewidth]{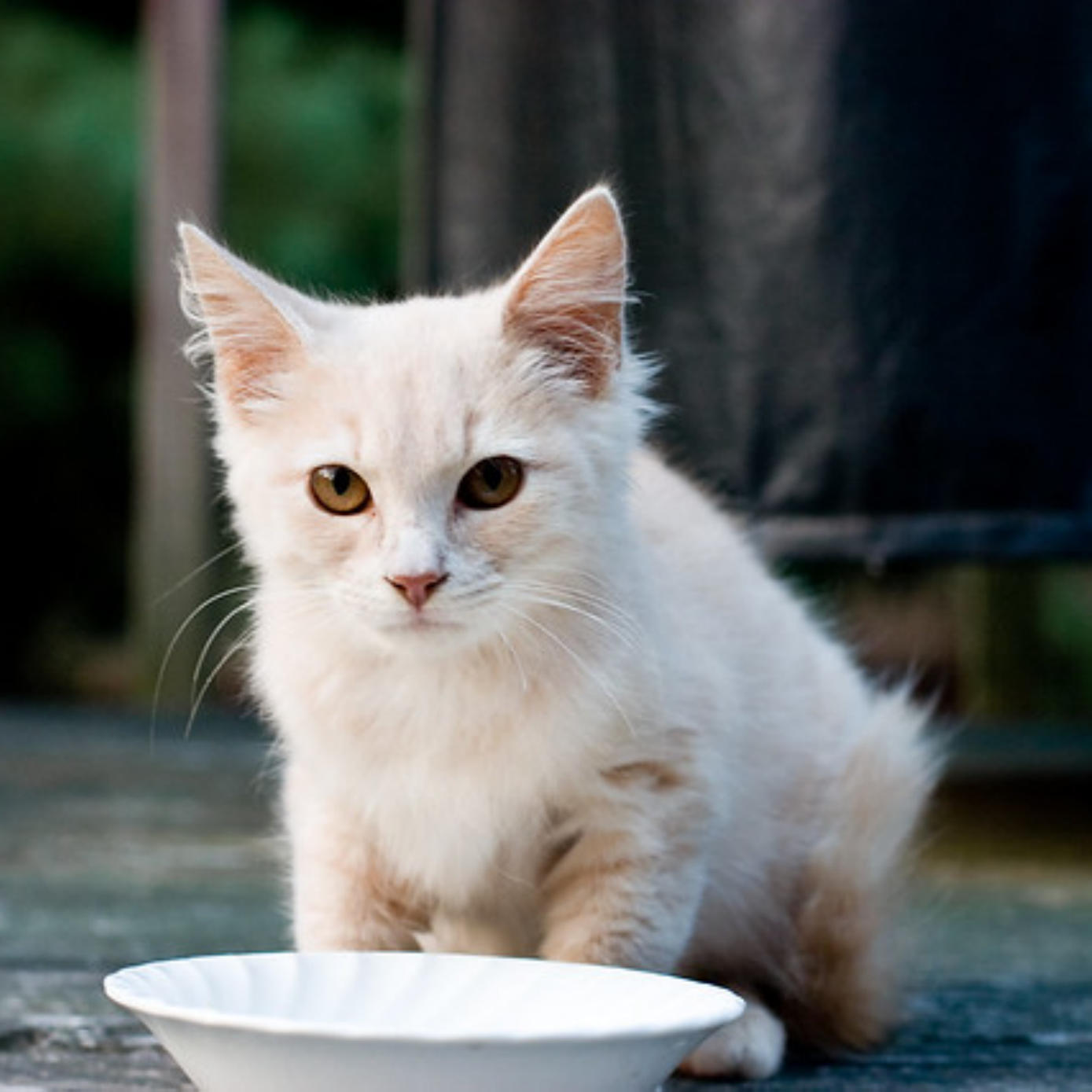}
	\end{minipage}
	}
	\subfigure[center heatmap]{
	\begin{minipage}[t]{0.22\linewidth} 
		\centering
		\includegraphics[page=2, width=\linewidth]{images/detection_heads.pdf}
	\end{minipage} 
	}
	\subfigure[width \& height]{
	\begin{minipage}[t]{0.22\linewidth} 
		\centering
		\includegraphics[page=3, width=\linewidth]{images/detection_heads.pdf}
	\end{minipage} 
    	}
	\subfigure[local shift]{
	\begin{minipage}[t]{0.22\linewidth}
		\centering
		\includegraphics[page=4, width=\linewidth]{images/detection_heads.pdf}
	\end{minipage}
	}
    \caption{The output heads of CenterNet for object detection.  See section \ref{section:system_design} for more details.}
    \label{fig:detection_heads}
\end{figure}

\subsection{\theSystem{} Design}
\label{section:system_design}
To exploit the full potential of hardware acceleration, we carefully select and integrate the operations and building blocks in \theSystem{}. 
We devise \theSystem{} to have the following embedded hardware compatible properties compared to other off-the-shelf network designs: 1) more uniform operation types to reduce the control complexity in the accelerator and to increase the accelerator utilization, 2) less computation to lower the overall latency to run on the embedded accelerator with limited compute capability, 3) smaller weights and inputs to be buffered on-chip for maximal reuse on the accelerator. 
Figure~\ref{fig:building_blocks} shows the basic building blocks as well as the overall network architecture of \theSystem{}. 

\textbf{Building Blocks and Feature Extractor}
The shaded part of Figure~\ref{fig:building_blocks} shows the basic building blocks of \theSystem{}.
Building block (a) is used to down-sample the input images. A 3$\times$3 depthwise convolution block with stride 2 is added to both of its branches together with 1$\times$1 convolution to aggregate information across the channel dimension.
Building block (b) splits the input features into two streams across the channel dimension. 
One branch is directly fed to the concatenation. 
The other streams through a sub-block of 1$\times$1, 3$\times$3 depthwise, and 1$\times$1 convolution. 
This technique is referred to as identity mapping~\cite{he2016identity}, 
which is commonly used to address the vanishing gradient problem during deep neural network training.
Building blocks (a) and (b) together form a shuffle block as shown in the left branch of the overall architecture in Figure~\ref{fig:building_blocks}, as part of the feature extractor ShuffleNetV2. 
We choose ShuffleNetV2 as it is one of the state-of-the-art efficient network design. ShuffleNetV2 1x configuration only requires 2.3M parameters (4.8$\times$ smaller than ResNet-18~\cite{he2016deep}) and 146M FLOPs of compute with resolution $224 \times 224$ (12.3x smaller than ResNet-18). Its top-1 accuracy is 69.4\% on ImageNet (0.36\% lower than ResNet-18). 

The deformable operation is used in building block (c).
Building block (c) is used to upsample the backbone features.
The first 1$\times$1 convolution is designed to map input channels to output channels.    
The following 3$\times$3 depthwise deformable convolution samples the previous feature map, according to the offsets generated by 1$\times$1 convolution.
After that, a $2\times$ upsampling layer, operated by a nearest neighbor kernel, is utilized to interpolate the higher resolution features. 
Note that, aside from the first layer, we only use 1$\times$1 convolution and 3$\times$3 depthwise (deformable) convolution in our build blocks. This way the building blocks of the whole network become more uniform and simple to support with specialized hardware.  

\textbf{Detection Heads}
As mentioned in Section \ref{section:2_1}, we use the anchor-free CenterNet~\cite{zhou2019objects} method to directly predict a gaussian distribution for object keypoints over the 2D space for object detection. 
Given an image $I \in \mathbb{R}^{W \times H \times 3}$, 
our feature extractor generates the final feature map $F \in \mathbb{R}^{\frac{W}{R} \times \frac{H}{R} \times D}$, 
where $R$ is the output stride and $D$ is the feature dimension. 
We set $R=4$ and $D=64$ for all the experiments.
As illustrated in Figure~\ref{fig:detection_heads}, the outputs include: 
\begin{enumerate}
    \item the keypoint heatmap $\hat Y \in [0, 1]^{\frac{W}{R} \times \frac{H}{R} \times C}$
    \item the object size $\hat S \in \mathbb{R}^{\frac{W}{R} \times \frac{H}{R} \times 2}$ 
    \item the local offset $\hat O \in \mathbb{R}^{\frac{W}{R} \times \frac{H}{R} \times 2}$
\end{enumerate}
Here $C$ is pre-defined as $20$ and $80$ for VOC and COCO, respectively.
In order to reduce the computation, we follow the class-agnostic practice, using the single size and offset predictions for all categories.
To construct bounding boxes from the keypoint prediction, we first collect the peaks in keypoint heatmap $\hat Y$ for each category independently. 
Then we only keep the top 100 responses which are greater than its eight-connected neighborhood.
Specifically, we use the keypoint values $\hat Y_{x_iy_ic}$ as the confidence measure of the $i$-th object for category $c$.
The corresponding bounding box is decoded as 
$ (\hat x_i + \delta \hat x_i - \hat w_i /2,\ \ \hat y_i + \delta \hat y_i - \hat h_i / 2, \hat x_i + \delta \hat x_i + \hat w_i / 2,\ \ \hat y_i + \delta \hat y_i + \hat h_i / 2)$, 
where $(\delta \hat x_i, \delta \hat y_i) = \hat O_{\hat x_i\hat y_i}$ is the offset prediction and $(\hat w_i,\hat h_i) = \hat S_{\hat x_i\hat y_i}$ is the size prediction.

\begin{figure*}[tp]
    \centering
    \includegraphics[width=0.98\linewidth]{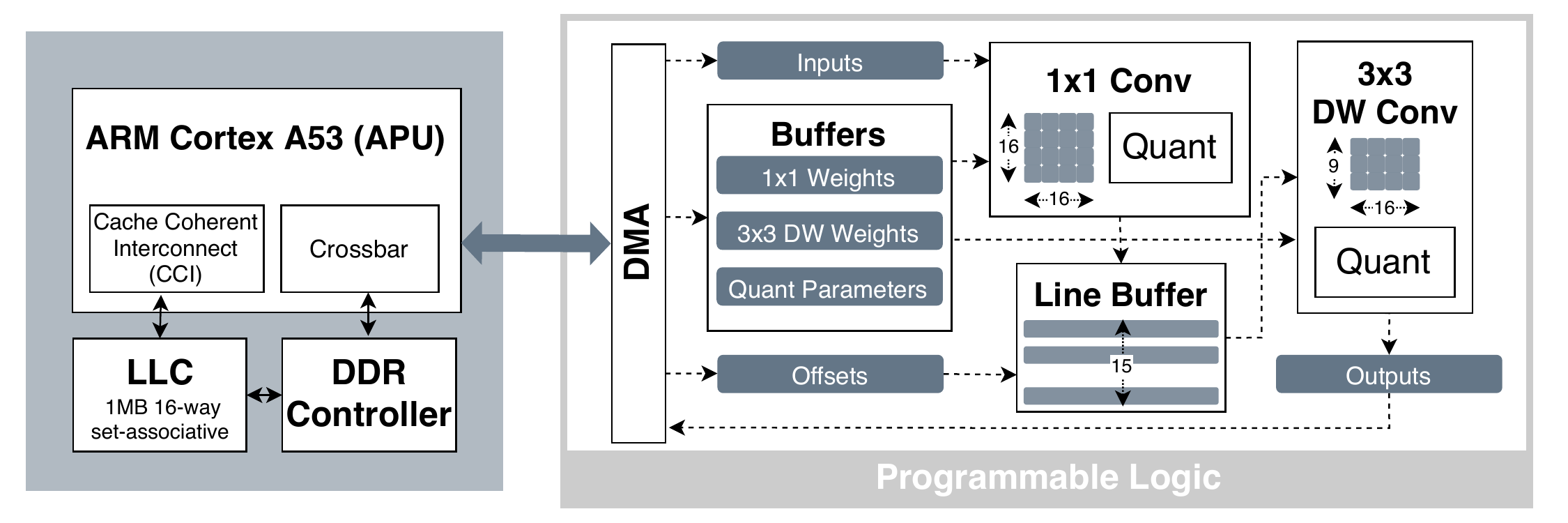} 
\caption{Architectural diagram of the FPGA accelerator.}
\label{fig:hw_arch}
\end{figure*} 

\textbf{Quantization}
Quantization is a crucial step towards the efficient deployment of the GPU pre-trained model on FPGA accelerators. Although many previous works treat quantization as a separate process outside the algorithm-hardware co-design loop, we note that quantization performance greatly depends on the network architecture. As an example, the residual connection will enlarge the activation range of specific layers, which makes a uniform quantization setting sub-optimal. And it requires a special design for addition in int32 format, otherwise, extra steps of quantization are needed to support the low-precision addition. With this prior knowledge, we use concatenation instead of residual connection throughout CoDeNet, and we do not use techniques such as layer aggregation~\cite{yu2018deep}, in order to achieve a simpler hardware design. 

We adopt a symmetric uniform quantizer shown as follows:
\begin{equation}
X^\prime = \text{clamp}(X, -t, t),
\end{equation}
\begin{equation}
X^I = \lfloor \frac{X^\prime}{\Delta} \rceil, \text{ where } \Delta = \frac{t}{2^{k-1}-1},
\end{equation}
\begin{equation}
Q(X) = \Delta X^I,
\end{equation}
where $Q$ stands for quantization operator, $X$ is a floating point input tensor (activations or weights), $\lfloor \cdot \rceil$ is the round operator, $\Delta$ is the quantization step (the distance between adjacent quantized points), $X^I$ is the integer representation of $X$, and $k$ is the quantization precision for a specific layer. Here, threshold value t determines the quantization range of the floating point tensor, and the clamp function sets all elements smaller than $-t$ to $-t$, and elements larger than $t$ to $t$. It should be noted that the threshold value t can be smaller than $max$ or $\mid$$min$$\mid$ in order to get rid of outliers and better represent the majority of a specific tensor. In order to achieve better AP, we perform 4-bit channel-wise quantization~\cite{krishnamoorthi2018whitepaper} for weights. Meanwhile, to ease the hardware design and accelerate the inference, we choose a symmetric uniform quantizer rather than non-uniform quantizer, and we use 8-bit layer-wise quantization for activations. During quantization-aware fine-tuning, we use Straight-Through Estimator (STE)~\cite{bengio2013estimating} to achieve the backpropagation of gradients through the discrete operation of quantization.

For the deformable convolution, quantization comprises two parts: 1) quantize the corresponding weights and activations, and 2) round and bound the sampling offsets of the deformable convolution. %
Compared to the standard convolution, the variable offsets will not significantly change the sensitivity of the network or the allowable quantization bit-width. 
Regarding the original fractional offsets, we bound and round them to be integers within the range $[-8, 7]$. This modification eliminates the need for bilinear interpolation and results in 1.9 AP drop on VOC as shown in Table~\ref{tab:algo_codesign}.

\begin{table*}[htp]
\caption{Quantized \theSystem{} on VOC object detection.}
    \centering
    \begin{tabular}{|l|c|c|c|c|c|c|c|c|c|}
    \hline
        \textbf{Detector} & \textbf{Resolution} & \textbf{DownSample} &\textbf{Weights} & \textbf{Activations} & \textbf{Model Size} & \textbf{MACs} & \textbf{AP50} \\ 
    \hline
        Tiny-YOLO & 416$\times$416 & MaxPool & 32-bit & 32-bit & 60.5 MB & 3.49 G & 57.1 \\
    \hline
        \multirow{2}{*}{\theSystem{}1$\times$ (config a)} & \multirow{2}{*}{256$\times$256} & \multirow{2}{*}{Stride4} & 32-bit & 32-bit & 6.06 MB & 0.29 G & 53.0 \\
         & &  & \cellcolor{Gray}4-bit & \cellcolor{Gray}8-bit & \cellcolor{Gray}0.76 MB & \cellcolor{Gray}0.29 G & \cellcolor{Gray}51.1 \\
        \hline
        \multirow{2}{*}{\theSystem{}1$\times$ (config b)}   & \multirow{2}{*}{256$\times$256} & \multirow{2}{*}{Stride2+MaxPool} & 32-bit & 32-bit & 6.06 MB & 0.29 G & 57.5 \\
         & & & \cellcolor{Gray}4-bit & \cellcolor{Gray}8-bit & \cellcolor{Gray}0.76 MB & \cellcolor{Gray}0.29 G & \cellcolor{Gray}55.1 \\
        \hline        
        \multirow{2}{*}{\theSystem{}1$\times$ (config c)}   &  \multirow{2}{*}{512$\times$512} & \multirow{2}{*}{Stride4} & 32-bit & 32-bit & 6.06 MB & 1.14 G & 64.6  \\
         & & & \cellcolor{Gray}4-bit & \cellcolor{Gray}8-bit & \cellcolor{Gray}0.76 MB & \cellcolor{Gray}1.14 G & \cellcolor{Gray}61.7 \\
    \hline
        \multirow{2}{*}{\theSystem{}2$\times$ (config d)}   &\multirow{2}{*}{512$\times$512} & \multirow{2}{*}{Stride4} & 32-bit & 32-bit & 23.2 MB & 3.54 G & 69.6 \\ 
        & & & \cellcolor{Gray}4-bit & \cellcolor{Gray}8-bit & \cellcolor{Gray}2.90 MB & \cellcolor{Gray}3.54 G & \cellcolor{Gray}67.1\\
        \hline        
        \multirow{2}{*}{\theSystem{}2$\times$ (config e)}  & \multirow{2}{*}{512$\times$512} & \multirow{2}{*}{Stride2+MaxPool} & 32-bit & 32-bit & 23.2 MB & 3.58 G & 72.4 \\ 
        & & & \cellcolor{Gray}4-bit & \cellcolor{Gray}8-bit & \cellcolor{Gray}2.90 MB & \cellcolor{Gray}3.58 G & \cellcolor{Gray}69.7\\
    \hline
    \end{tabular}

\label{tab:algo_codesign_voc_quantization}
\end{table*}

\begin{table*}[htp]
    \centering
\caption{Quantized \theSystem{} on COCO object detection.}    
    \begin{tabular}{|c|c|c|c|c|c|c|c|c|c|}
    \hline
        \textbf{Detector} & \textbf{Weights} & \textbf{Model Size} & \textbf{MACs} & \textbf{AP} & \textbf{AP50} & \textbf{AP75} & \textbf{APs} & \textbf{APm} & \textbf{APl} \\ 
    \hline 
        \multirow{2}{*}{\theSystem{}1$\times$}  & 32-bit & 6.07MB & 1.24G & 22.2 & 38.3 & 22.4 & 5.6 & 22.3 & 38.0 \\
         & 4-bit  & 0.76MB & 1.24G & 18.8 & 33.9 & 18.7 & 4.6 & 19.2 & 32.2 \\
    \hline
        \multirow{2}{*}{\theSystem{}2$\times$} & 32-bit & 23.4MB & 4.41G & 26.1 & 43.3 & 26.8 & 7.0 & 27.9 & 43.5 \\
         & 4-bit & 2.93MB & 4.41G & 21.0 & 36.7 & 21.0 & 5.8 & 22.5 & 35.7 \\
    \hline
    \end{tabular}
\label{tab:algo_codesign_coco_quantization}
\end{table*}

\subsection{Dataflow Accelerator}
We develop a specialized accelerator to support the aforementioned \theSystem{} design on an FPGA SoC. 
As shown in Figure~\ref{fig:hw_arch}, the FPGA SoC includes the programmable logic (PL), memory interfaces, a quad-core ARM Cortex-A53 application processor with 1MB LLC, and etc. 
Our accelerator on the PL side communicates to the processor through an AXI system bus.
The High Performance (HP) and Accelerator Coherency Port (ACP) interfaces on the AXI bus allow the accelerator to directly access the DRAM or perform cache-coherent accesses to the LLC and DRAM. 
The processor provides software support to invoke the accelerator and to run functions that are not implemented on the accelerator. 

With our co-design methodology, we are able to reduce the types of operations to support in the accelerator. Excluding the first layer for the full $3\times3$ convolution, \theSystem{} only consists of the following operations: (i) $1\times1$ convolution, (ii) $3\times3$ depthwise (deformable) convolution, (iii) quantization, (iv) split, shuffle and concatenation. 
This helps us simplify the complexity of the control logic and thus saves more FPGA resources for the actual computation.
We partition the \theSystem{} workload so that the frequently-called compute-intensive operations are offloaded to the FPGA accelerator while the other operations are run by software on the processor. 
The operations we choose to accelerate are $1\times1$ convolution, $3\times3$ depthwise (deformable) convolution, and quantization, with the other operations offloaded to the processor. 

To leverage both the data-level and the task-level parallelism, we devise a spatial dataflow accelerator engine to execute a subgraph of the \theSystem{} at a time and store the intermediate outputs to the DRAM. 
In the dataflow engine, the execution of compute units is determined by the arrival of the data and thus further reduces the overhead from the control logic. 
As illustrated in the architectural diagram in Figure~\ref{fig:hw_arch}, our accelerator executes $1\times1$ convolution with quantization and $3\times3$ depthwise (deformable) convolution with quantization in order. We implement the accelerator with Vivado HLS and its dataflow template. All functional engines are connected to each other through data FIFOs. 
Extra bypass signals can be asserted if the user would like to bypass either of the main computation blocks. By co-designing the network to use operations with fewer weight parameters, such as depthwise convolution, we are able to buffer the weights for all operations in the on-chip memory and enable the maximal reuse of the weights once they are on-chip. 
We also add a line buffer for the $3\times3$ depthwise (deformable) convolution to maximize the reuse of inputs on-chip. This optimization is enabled by the operation co-design discussed in Section~\ref{section:3_2}. 
The line buffer stores 15 rows of the input image. The size of this buffer is larger than $15 \times w \times ic$ of any layers in the \theSystem{} design. 
Our input tensors are laid out in the NHWC manner, allowing the data along the channel dimension C to be stored in contiguous memory blocks.

\textbf{$1\times1$ convolution} The compute engine for the $1\times1$ convolution is composed of $16 \times 16$
multiply-accumulate (MAC) units.  At each round of the run, the engine takes 16 inputs along its channel dimension and broadcasts each of them to 16 MAC units. Meanwhile, it unicasts $16 \times 16$ weights for 16 input channels and 16 output channels to their corresponding MAC unit. There are 16 reduction trees of size 16 connected with the MAC units to generate 16 partial sums of the products. The partial sums are stored on the output registers and are accumulated across each round of the run. Every time the engine finishes the reduction along the input channel dimension, it feeds the values of the output registers to the output FIFO and resets their values to zero. 

\textbf{$3\times3$ depthwise (deformable) convolution}
This engine directly reads 16 sampled $3 \times 3$ inputs from the line buffer design and multiplies them by $3 \times 3$ weights from 16 corresponding channels. Then it computes the outputs with 16 reduction trees to accumulate the partial sums along $3 \times 3$ spatial dimension. Both the original and the deformable depthwise convolutions can be run on this engine. The original depthwise operation is realized by hardcoding the offset displacement to be 1.

\textbf{Quantization} 
To convert the output from the 16-bit sum to 8-bit inputs, we add a quantization unit at the end of each compute engine. The quantization unit multiplies each output with a scale, and then add a bias to it. 
It returns the lower 8 bits of the result as the quantized value. 
The parameters, such as the scale and bias for each channel, are preloaded to the on-chip buffer to save the memory access time. 
Note that we also merge the batch normalization and ReLU in this compute unit.  
We follow the practice introduced in \cite{jacob2018quantization} to perform integer inference for our quantized model.

Our accelerator design can execute
$16\times1\times250\times2=128$ GOPs for 1$\times$1 convolution and $9\times16\times250\times2=72$ GOPs for 3$\times$3 depthwise convolution simultaneously. 
On our target FPGA with 6GB/s DDR bandwidth, we can load 4 Giga pairs of 8-bit inputs and 4-bit weights per second. The arithmetic intensity required to reach the compute bound is $128/4=32$ OPs/pair for 1$\times$1 convolution and $72/4=18$ OPs/pair for 3$\times$3 depthwise convolution. Our buffering strategy allows us to reach the compute bound through the reuse of weights and the activations.

\section{Experimental Results}
\label{section:result}
\begin{table*}[htp]
    \centering
\caption{Performance comparison with prior works.}    
\begin{tabular}{|l|l|l|l|l|l|r|}
\hline
\textbf{}   & \textbf{Platform} & \textbf{Input Resolution} & \textbf{Framerate (fps)} & \textbf{Test Dataset} & \textbf{Precision} & \textbf{Accuracy} \\ \hline
DNN1~\cite{hao2019fpga} & Pynq-Z1 & - & 17.4  & \multirow{3}{*}{DJI-UAV} & a8  & IoU(68.8)  \\ %
DNN3~\cite{hao2019fpga} & Pynq-Z1 & - & 29.7  &   & a16  & IoU(59.3) \\
Skynet~\cite{zhang2019skynet}  & Ultra96 & 160 $\times$ 360 & 25.5  &  & w11a9 &  IoU(71.6)  \\ \hline %
AP2D~\cite{li2020novel} & Ultra96 &  224 $\times$ 224 & 30.5 & AD2P & w(1-24)a3 & IoU(55)\\ \hline

Finn-R~\cite{blott2018finn}~\cite{preusser2018inference}  & Ultra96 & - & 16 & \multirow{2}{*}{VOC07}  & w1a3  & AP50(50.1) \\ %
Tiny-Yolo-v2~\cite{farrukh2020power} & Zynq-706 XC7Z045 & 224 $\times$ 224 & 43.1 & & w16a16 & AP50(48.5)\\

\rowcolor{Gray}
\textbf{Ours (config a)}&  & 256 $\times$ 256 & 32.2 &  &  & AP50(51.1) \\
\rowcolor{Gray}
\textbf{Ours (config b)}&  & 256 $\times$ 256 & 26.9 &  &  & AP50(55.1) \\
\rowcolor{Gray}
\textbf{Ours (config c)}& Ultra96 & 512 $\times$ 512 & 9.3 & VOC07 & w4a8 & AP50(61.7) \\ 
\rowcolor{Gray}
\textbf{Ours (config d)}&  & 512 $\times$ 512 & 5.2 &  &  & AP50(67.1) \\ 
\rowcolor{Gray}
\textbf{Ours (config e)}&  &512 $\times$ 512 & 4.6 &&  & AP50(69.7) \\ 
\hline
\end{tabular}
\label{tab:time}
\end{table*} 
We implement \theSystem{} in PyTorch, train it with a pretrained ShuffleNetV2 backbone, and quantize the network to use 8-bit activations and 4-bit weights. 
We devise several configurations of \theSystem{}
to facilitate the latency-accuracy tradeoffs for our final object detection solution on the embedded FPGAs.
Different configurations of the \theSystem{} are listed in Table~\ref{tab:algo_codesign_voc_quantization} and \ref{tab:algo_codesign_coco_quantization} showing the accuracies for object detection on Pascal VOC and Microsoft COCO 2017 dataset. 

In Table~\ref{tab:algo_codesign_voc_quantization}, we show different configurations of \theSystem{} with an accuracy-efficiency trade-off. \emph{config c, d} and \emph{e} use image size $512\times512$, which is the default resolution of CenterNet. Compared to Tiny-YOLO, our \emph{config c} model is $10\times$ smaller without quantization and $79.6\times$ smaller with quantization, while achieving higher accuracy. In addition, the total MACs count of our compact design is $3.1\times$ smaller than Tiny-YOLO. It can be seen that quantizing the model to 4--8 bits causes a minor accuracy drop, but can significantly reduce the model size ($> 8\times$). In order to further save the MACs, we reduce the resolution to be $256\times256$, corresponding to \emph{config a}, where we can still get 53 AP50 with about 1/4 total MACs compared with \emph{config c}. Moreover, we found the downsampling strategy of the first layer play an important role, where a larger stride can benefit the speed (shown later in Table~\ref{tab:time}), but a smaller stride processes more information and can therefore improve accuracy (corresponding to \emph{config b}). For scenarios that require more accurate detectors, we expand the channel size of \emph{config c} (\theSystem{}1$\times$) by a factor of 2, which gives us config d that can achieve 69.6 AP50. After quantization, \emph{config d} has a 67.1 AP50 with comparable MACs but 21$\times$ smaller memory size compared to Tiny-YOLO. By doubling the channel size (\theSystem{}2$\times$) and using a smaller stride, we have \emph{config e}, which can achieve the highest 72.4 AP50 among all the configurations.

Table~\ref{tab:algo_codesign_coco_quantization} shows the accuracy of \theSystem{}s on the Microsoft COCO 2017 dataset. Microsoft COCO is a more challenging dataset compared to Pascal VOC, where COCO has 80 categories but Pascal VOC has 20. Our results here are obtained with default $512\times512$ resolution, and with stride 2 convolution and maxpooling as the downsampling strategy. Besides AP50, COCO primarily uses AP as the evaluation metric, which is the average among AP[0.5:0.95] (namely AP50, AP55, ..., AP95). As we can see in the table, \theSystem{}1$\times$ can achieve 22.2 AP with model size 6.07 MB. Applying quantization will cause a minor accuracy degradation, but can get an 8$\times$ smaller model. The same trend holds for \theSystem{}2$\times$ where our model can get 26.1 and 21.0 AP, with and without quantization respectively.

We evaluate our accelerator customized for each \theSystem{} configurations on the Ultra96 development board with Xilinx Zynq XCZU3EG UltraScale+ MPSoC device. 
Our accelerator design runs at 250 MHz after synthesis, and place and route. 
Table \ref{tab:area} shows the overall resource utilization of our implementation. %
We observe a 100\% utilization of both DSPs and BRAMs. Most DSPs are mapped to the 4-8 bit MAC units, and BRAMs are mainly used for the line buffer design. 
Our Power measurements are obtained via a power monitor. We measured 4.3W on the Ultra96 power supply line  with no workload running on the programming logic side and 5.6W power when running our network. On \theSystem{} \emph{config a}, our accelerator achieves 5.75 fps / W in terms of power efficiency.

\begin{figure}[tp]
\vspace{3mm}
\begin{tikzpicture}
	\begin{axis}[height=6cm, width=0.5\textwidth,
		xlabel={Accelerator Inference Time (ms)},
		ylabel={VOC AP50}]
	\addplot[color=teal,mark=*,] coordinates {

(219, 69.7)
(194, 67.1)
(108, 61.7)
(37, 55.1)
(31, 51.1)
	};
	\node [below, color=teal] at (axis cs:  219, 69.7) {e};
	\node [below, color=teal] at (axis cs:  194, 67.1) {d};
	\node [above, color=teal] at (axis cs:  108, 61.7) {c};	
	\node [above, color=teal] at (axis cs:  37, 55.1) {b};
	\node [above, color=teal] at (axis cs: 31, 52.0) {a};
	
	\addplot[color=red, mark=square*,]
	coordinates {
(62.5, 50.1)
	};
	\node [right, color=red] at (axis cs: 62.5, 50.1) {\ \ \ FINN-R};%
	
	\addplot[color=red, mark=square*,]
	coordinates {
	(23.2, 48.5)
	};
	\node [right, color=red] at (axis cs: 23.2, 48.5) {\ \ \ Tiny-Yolo-v2};%
	\end{axis}
\end{tikzpicture}
\vspace{-3pt}
\caption{Latency-accuracy trade-off on VOC.}
\label{fig:tradeoff}
\vspace{-4pt}
\end{figure}
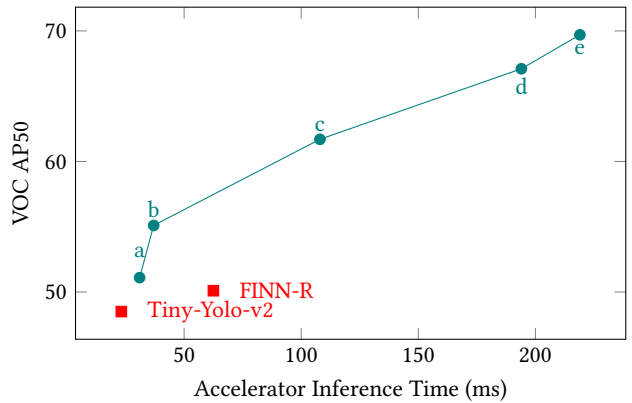

We provide a pareto curve in Figure~\ref{fig:tradeoff} showing the latency-accuracy tradeoff for various \theSystem{} design points with acceleration. Configuration $a$ and $b$ in this curve are trained and inferenced with images of size $256\times256$ instead of the original size $512\times512$. The smaller input image size leads to $\sim$4$\times$ reduction in MACs. In configuration $a$, $c$ and $d$, the stride of the first layer is increased from 2 to 4, which greatly reduces the first layer runtime on the processor. In configuration $d$ and $e$, we use the \theSystem{} 2$\times$ model, where the channel size is doubled in the network, to boost the accuracy.  
The latency evaluation on our accelerator is done with the batch size equal to 1 without any runtime parallelization. We run the first layer of the network on the processor for all configurations.  

\begin{table}[tp]
\centering
\caption{FPGA resource utilization.}
\begin{tabular}{|c|c|c|c|c|}
\hline
\textbf{LUT} &\textbf{FF} & \textbf{BRAM} & \textbf{DSP} \\
\hline
34144 (48.4\%) & 41827 (29.6\%) & 216 (100\%) & 360 (100\%) \\ \hline
\end{tabular}
\label{tab:area}
\end{table}

A comparison of our solutions against previous works is shown in Table~\ref{tab:time}.
We found that very few prior works on embedded FPGAs attempt to target the standard dataset like VOC or COCO for object detection, primarily due to the challenges from limited hardware resources and inefficient model design. 
Two state-of-the-art FPGA solutions that meet the real-time requirement in the DAC-UAV competition target the DJI-UAV dataset for drone image detection. 
However, object detection on DJI-UAV is a less generic and less challenging task than object detection on VOC or COCO. The images in DJI-UAV dataset are taken from the top-down view. They typically contain very few overlapped objects. 
In addition, the DJI-UAV dataset is designed for single-object detection whereas VOC and COCO can be used for multi-object detection. 
Hence, in this work, we target VOC and COCO to provide a more general solution for multi-object detection and for images taken from the most common first-person view. %

As shown in Figure~\ref{fig:tradeoff} and Table~\ref{tab:time}, compared to the results from FINN-R~\cite{blott2018finn}~\cite{preusser2018inference}, the state-of-the-art embedded FPGA accelerator design targeting VOC, our configuration $a$ and $b$ (with single-batch inference latency of 31ms and 37ms respectively) achieve both higher accuracy, higher framerate, and lower latency. Another state-of-the-art work Tiny-Yolo-v2~\cite{farrukh2020power} attains low latency, but with lower accuracy. It also runs on a different FPGA platform.

\section{Conclusion}
\label{section:conclusion}
In this work, we performed a detailed accuracy-efficiency trade-off study for each hardware-friendly algorithmic modification to the input-adaptive deformable convolution operation, with the goal of co-designing an efficient object detection network and a real-time embedded accelerator optimizing for accuracy, speed, and energy efficiency.
Results show that these modifications led to significant hardware performance improvement on the accelerator with minor accuracy loss. Our co-designed model \theSystem{} with the modified deformable convolution is $79.6\times$ smaller than Tiny YOLO and its corresponding embedded FPGA accelerator is able to achieve real-time processing with a framerate of 26.9. 
Our higher-accuracy \theSystem{} model achieves to 67.1 AP50 on Pascal VOC with only 2.9 MB of parameters---$20.9\times$ smaller but 10\% more accurate than Tiny-YOLO.

\section*{Acknowledgements}
This work was supported in part 
by the CONIX Research Center, one of six centers in JUMP,
a Semiconductor Research Corporation (SRC) program sponsored by DARPA,
and by Facebook Reality Labs, Google Cloud, Samsung SAIT, 
by the Berkeley ADEPT Lab, Berkeley Deep Drive, and the Berkeley Wireless Research Center.
\clearpage

\bibliographystyle{ACM-Reference-Format}
\bibliography{ref}
\end{document}